\documentclass[conference]{IEEEtran}
\IEEEoverridecommandlockouts
\usepackage{cite}
\usepackage{amsmath,amssymb,amsfonts}
\usepackage{algorithmic}
\usepackage{graphicx}
\usepackage{textcomp}
\usepackage{xcolor}
\def\BibTeX{{\rm B\kern-.05em{\sc i\kern-.025em b}\kern-.08em T\kern-.1667em\lower.7ex\hbox{E}\kern-.125emX}}
\allowdisplaybreaks
\DeclareMathOperator{\tr}{tr}
\DeclareMathOperator{\diam}{diam}
\DeclareMathOperator{\sign}{sign}
\newcommand{\E}{\mathop{{}\mathbb E}} 
\newcommand{\I}[1]{\mathbb I\{#1\}} 
\newtheorem{theorem}{Theorem}
\newtheorem{definition}[theorem]{Definition}
\newtheorem{lemma}[theorem]{Lemma}
\newtheorem{remark}[theorem]{Remark}

\def\BibTeX{{\rm B\kern-.05em{\sc i\kern-.025em b}\kern-.08em
    T\kern-.1667em\lower.7ex\hbox{E}\kern-.125emX}}
\begin{document}

\title{Compressive Mahalanobis Metric Learning Adapts to Intrinsic Dimension}

\author{\IEEEauthorblockN{Efstratios Palias}
\IEEEauthorblockA{\textit{School of Computer Science} \\
\textit{University of Birmingham}\\
Birmingham, United Kingdom \\
exp093@bham.ac.uk}
\and
\IEEEauthorblockN{Ata Kab\'an}
\IEEEauthorblockA{\textit{School of Computer Science} \\
\textit{University of Birmingham}\\
Birmingham, United Kingdom \\
a.kaban@bham.ac.uk}
}

\maketitle

\begin{abstract}
Metric learning aims at finding a suitable distance metric over the input space, to improve the performance of distance-based learning algorithms. In high-dimensional settings, it can also serve as dimensionality reduction by imposing a low-rank restriction to the learnt metric. In this paper, we consider the problem of learning a Mahalanobis metric, and instead of training a low-rank metric on high-dimensional data, we use a randomly compressed version of the data to train a full-rank metric in this reduced feature space. We give theoretical guarantees on the error for Mahalanobis metric learning, which depend on the stable dimension of the data support, but not on the ambient dimension. Our bounds make no assumptions aside from i.i.d. data sampling from a bounded support, and automatically tighten when benign geometrical structures are present. An important ingredient is an extension of Gordon's theorem, which may be of independent interest. We also corroborate our findings by numerical experiments.
\end{abstract}


\section{Introduction}
In clustering and classification, there have been numerous distance-based algorithms proposed. While the Euclidean metric is the ``standard'' notion of distance between numerical vectors, it does not always result in accurate learning. This can be e.g. due to the presence of many dependent features, noise, or features with large ranges that dominate the distances \cite{verma2015sample}. Mahalanobis metric learning aims at lessening this caveat by linearly transforming the feature space in a way that properly weights all important features, and discards redundant ones. In its most common form, metric learning focuses on learning a Mahalanobis metric \cite{xing2002distance,weinberger2009distance,davis2007information}.

Metric learning algorithms can be divided into two types based on their purpose \cite{verma2015sample}. \emph{Distance-based metric learning} aims to increase the distances between instances of different classes (inter-class distances) and decrease the distances inside the same class (intra-class distances). On the other hand, \emph{classifier-based metric learning} focuses on directly improving the performance of a particular classification algorithm, and is therefore dependent on the algorithm in question.

Despite the success of Mahalanobis metric learning, high-dimensionality of the data is a provable bottleneck that arises fairly often in practice. The work of \cite{verma2015sample} has shown, through both upper and lower bounds, that, in the absence of assumptions or constraints, the sample complexity of Mahalanobis metric learning, increases linearly with the data dimension. In addition, so does the computational complexity of learning. Compounding this, high-dimensionality is known to quickly degrade the performance of machine learning algorithms in practice. This means that, even if a suitable distance metric is found, the subsequent algorithm might still perform poorly. All these issues are collectively known as the \emph{curse of dimensionality} \cite{verleysen2005curse}.

It has been observed, however, that many real-world data sets do not fill their ambient spaces evenly in all directions, but instead their vectors cluster along a low-dimensional subspace with less mass in some directions, or have many redundant features \cite{pope2020intrinsic}. We refer to these data sets, in a general sense, in a broad sense, as having a low \emph{intrinsic dimension} (low-ID). Due to their lower information content, it is intuitively expected that learning from such a data set should be easier, both statistically and computationally. One of the most popular ways to take advantage of a low-ID is to compress the original data set into a low-dimensional space \cite{achlioptas2003database} and then proceed with learning in this smaller space \cite{KR24}.

Random projections is a widely used compression method with attractive theoretical guarantees. These are universal in the sense of being oblivious to the data being compressed. All instances are subjected to a random linear mapping without significantly distorting Euclidean distances, and reducing subsequent computing time. There has been much research on controlling the loss of accuracy with random projections for various learning algorithms, see e.g. \cite{reboredo2016bounds,li2019random}. Another advantage, is that no pre-processing step is necessary beforehand, making random projections simple to implement \cite{achlioptas2003database}. In the case of Mahalanobis metric learning, an additional motivation is to reduce the number of parameters to be estimated.

\subsection{Our contributions}\label{subsec:contributions}
We consider the problem of learning a Mahalanobis metric from random projections (RP) of the data, and for the case of Gaussian RP give the following theoretical guarantees:
\begin{itemize}
    \item a high-probability uniform upper bound on the generalisation error 
    \item a high-probability upper bound on the excess empirical error of the learnt metric, relative to the empirical error of the metric learnt in the original space.
\end{itemize}

The quantities in these two theoretical guarantees (given in Theorems \ref{th:generalisation error} and \ref{th:excess empirical error} respectively) 
capture a trade-off in compressive learning of a Mahalanobis metric: as the projection dimension decreases the first quantity becomes lower and the second becomes higher.

Most importantly, unlike metric learning in the original high-dimensional space, we find that neither of these two quantities depend on the ambient dimension explicitly, but only through a notion of ID, namely the so-called \emph{stable dimension}, defined in Definition \ref{def:stable dimension}.
This shows that the aforementioned trade-off can be reduced, should the stable dimension be low. We corroborate our theoretical findings with numerical experiments on synthetic data in order to show the extent to which the stable dimension plays a role in the effectiveness of metric learning in practice.

As an important ingredient of our analysis, we revisit a well-known result due to Gordon \cite{gordon1988milman} that uniformly bounds the maximum norm of vectors in the compressed unit sphere under a Gaussian RP. We extend this result into a dimension-free version, for arbitrary domains, in Lemma \ref{th:Gordon's theorem generalisation}, which may be of independent interest.


\subsection{Related work}\label{subsec:related work}
Mahalanobis metric learning was introduced in \cite{xing2002distance} and has attracted a significant amount of research since. Shortly after its introduction, two of the most popular metric learning algorithms were proposed; Large Margin Nearest Neighbour (LMNN) \cite{weinberger2009distance}, and Information Theoretic Metric Learning (ITML) \cite{davis2007information}. Generalisations and extensions to metric learning algorithms have also been well-studied. We refer the reader to the surveys in \cite{kulis2013metric,wang2015survey} for a more detailed review on metric learning algorithms. There have also been attempts to learn non-linear metrics (e.g. \cite{kedem2012non,chen2019curvilinear}), as well as to train neural networks in metric learning, known as {deep metric learning} (see \cite{kaya2019deep} for a survey). Metric learning has also been applied to other fields, e.g. collaborative filtering \cite{hsieh2017collaborative}, and facial recognition \cite{guillaumin2009you}.

Much recent literature has been devoted to mitigate the undesirable effects of the curse of dimensionality on metric learning. A typical approach is to train a low-rank metric in the ambient space, this was demonstrated to improve the classification performance -- see e.g. \cite{xie2018orthogonality} and the references therein. In
%
\cite{verma2015sample}, the authors consider both distance-based and classifier-based Mahalanobis metric learning, and 
show that sample complexity necessarily grows with the number of features, unless a Frobenius norm-constraint is imposed onto the hypothesis class of Mahalanobis metrics. 
In a closely related model, namely a quadratic classifier class, it was found in \cite{latorre2021effect} that a nuclear-norm constraint leads to the ability of the error to automatically adapt to a notion of intrinsic dimension of the data (the effective rank of the true covariance), while the Frobenius norm constraint was shown to lack such ability. Their bound still has a mild logarithmic dependence on the ambient dimension.

All of the above methods (and most others) work with the full data set, which can be limiting with high-dimensional data. 
Novel data acquisition sensors from compressed sensing enable collecting data in a randomly compressed form, alleviating the need to select and discard significant fractions of it during pre-processing \cite{yazicigil2019taking}.
\section{Theoretical results}\label{sec:theoretical results}
\textbf{Notation}: We denote scalars and vectors by lowercase letters, and matrices with capital letters. The Euclidean norm of a vector is denoted $\|\cdot\|$, whereas the Frobenius norm of a matrix is denoted $\|\cdot\|_F$. The trace of a matrix is denoted $\tr(\cdot)$. We let $\sigma_{\min}(\cdot)$ and $\sigma_{\max}(\cdot)$ be respectively the smallest and largest singular values of a matrix. $I_n$ denotes the $n\times n$ identity matrix, and $0_n$ denotes the $n$-dimensional zero vector. The notation $\mathcal N(\mu,\Sigma)$ stands for the Gaussian distribution with mean vector $\mu$ and covariance matrix $\Sigma$. We denote by $\E\cdot$ (without brackets) the expectation of a random variable (or random vector). $\I{\cdot}$ is the indicator function, that equals $1$ if its argument is true, and $0$ otherwise. We denote by $\mathcal S^{n-1}$ the $n$-dimensional unit sphere. For a set $T$, we write $\diam(T):=\sup_{x,x'\in T}\|x-x'\|$ for its diameter, and $T-T:=\{x-x':x,x'\in T\}$. With a slight abuse of notation, if $T$ is a set and $A$ is a conformable matrix, we write $AT:=\{Ax:x\in T\}$.

We now formally introduce the problem of Mahalanobis metric learning, as well as the random compression that we use. Let $\mathcal X\times\mathcal Y$ be the instance space, where $\mathcal X\subset\mathbb R^d$ is the feature space and $\mathcal Y=\{0,1\}$ is the set of labels. We consider the usual setting where all instances are assumed to have been sampled i.i.d. from a fixed but unknown distribution $\mathcal D$ over $\mathcal X\times\mathcal Y$. For our derivations, the diameter of $\mathcal X$ is assumed finite, that is $\diam(\mathcal X)<\infty$.

The goal of Mahalanobis metric learning is to learn a matrix $M\in\mathbb R^{d\times d}$, such that the Mahalanobis distance between any two instances $x,x'$, i.e. $\|Mx-Mx'\|$, is larger if $x,x'$ have different labels and smaller if $x,x'$ share the same label. For the purpose of dimensionality reduction, given a fixed $k$, where $k\leq d$, we let $R\in\mathbb R^{k\times d}$ be our random projection (RP) matrix. We assume that each datum instance is available only in its RP-ed form (as in compressed sensing applications). We will be referring to $d$ and $k$ as the \emph{ambient dimension} and the \emph{projection dimension} respectively.

While there are several possible choices for the random matrix $R$, in our theoretical analysis we employ the \emph{Gaussian random projection}. That is, the elements of $R$ are drawn i.i.d. from $\mathcal N(0,1/k)$. The motivation for this choice is twofold: it is known to have the ability to approximately preserve the relative distances among the original data with high probability \cite{indyk1998approximate,dasgupta2003elementary}, and it also allows us to employ some specialised theoretical results for tighter guarantees.

Next, we define the hypothesis classes of Mahalanobis metrics. Let
\begin{equation}
    \mathcal M:=\{M_0\in\mathbb R^{d\times d}:\sigma_{\max}(M_0)=1/\diam(\mathcal X)\}
\label{eq:metric class}\end{equation}
be the hypothesis class in the ambient space, and
\begin{equation}
    \mathcal M_k:=\{M\in\mathbb R^{k\times k}:\sigma_{\max}(M)=1/\diam(\mathcal X)\}
\label{eq:compressed metric class}\end{equation}
be the hypothesis class in the compressed space $R\mathcal X$, where the constraints on $\sigma_{\max}$ are to avoid arbitrary scaling, and to make our main results scale-invariant. Let
\begin{equation}
    T:=\{((x_{2i-1},y_{2i-1}),(x_{2i},y_{2i}))\}_{i=1}^n 
\label{eq:training set}\end{equation}
be a training set of $n$ pairs of instances from $\mathcal X\times\mathcal Y$. Also, let $\ell_{l,u}:\mathbb R\times\{0,1\}\to[0,1]$ be a distance-based loss function defined as
\begin{equation}
    \ell_{l,u}(x,y):=\begin{cases}
        \min\{1,\rho(x-u)_+\}\text{ if }y=1\\
        \min\{1,\rho(l-x)_+\}\text{ if }y=0.
    \end{cases}
\label{eq:loss function}\end{equation}
where $(\cdot)_+:=\max\{\cdot,0\}$, and $\rho,l,u$ are positive numbers with $l<u$.

Note that $\ell_{l,u}$ is $\rho$-Lipschitz in its first argument, a property we exploit later in the derivations. This loss function penalizes small inter-class distances and large intra-class distances, and is a common choice for distance-based metric learning \cite{verma2015sample}.

We next define the true error of a hypothesis $M\in\mathcal M_k$, given the matrix $R$, as
\begin{equation}\label{eq:true error compressed}
    L^R_{\mathcal D}(M):=\E_{((x,y),(x',y'))\sim\mathcal D^2}\ell_{l,u}(\|MRx-MRx'\|^2,\I{y=y'}),
\end{equation}
and its empirical error, given the training set $T$ from \eqref{eq:training set}, as
\begin{equation}\label{eq:empirical error compressed}
    \hat L^R_T(M):=\frac1n\sum_{i=1}^n\ell_{l,u}(\|MRx_{2i-1}-MRx_{2i}\|^2,\I{y_{2i-1}=y_{2i}}).
\end{equation}

For a hypothesis $M_0\in\mathcal M$, the true and empirical error are defined analogously, by omitting $R$ and considering the original vectors. That is, the true error is defined as
\begin{align}\label{eq:true error}
    L_{\mathcal D}(M_0):=\E_{((x,y),(x',y'))\sim\mathcal D^2}\ell_{l,u}(\|M_0x-M_0x'\|^2,\I{y=y'}),
\end{align}
and the empirical error is defined as
\begin{align}\label{eq:empirical error}
    \hat L_T(M_0):=\frac1n\sum_{i=1}^n\ell_{l,u}(\|M_0x_{2i-1}-M_0x_{2i}\|^2,\I{y_{2i-1}=y_{2i}}).
\end{align}

We would first like to upper bound the generalisation error $(L^R_{\mathcal D}(M)-\hat L^R_T(M))$, uniformly, for all $M\in\mathcal M_k$, with high-probability, with respect to the random draws of $R$. To this end, let us introduce some complementary definitions and results, that appear in the derivations.
\begin{definition}[Gaussian width {\cite[Definition 7.5.1]{vershynin2018high}}]\label{def:Gaussian width}
    Let $T\subset\mathbb R^d$ be a set, and $g\sim\mathcal N(0_d,I_d)$. The \emph{Gaussian width} of $T$, is defined as
    \begin{equation}
        \omega(T):=\E\sup_{x\in T}g^\top x,
    \end{equation}
    and the \emph{squared version} of the Gaussian width of $T$, is defined as
    \begin{equation}
        \psi(T):=\sqrt{\E\sup_{x\in T}(g^\top x)^2}.
    \end{equation}
\end{definition}

Definition \ref{def:Gaussian width} allows us to introduce a more robust version of the algebraic dimension, as follows.

\begin{definition}[Stable dimension {\cite[Definition 7.6.2]{vershynin2018high}}]\label{def:stable dimension}
    The \emph{stable dimension} of a set $T\subset\mathbb R^d$, with $0<\diam(T)<\infty$, is defined as
    \begin{equation}
        s(T):=\frac{\psi(T-T)^2}{\diam(T)^2}.
    \end{equation}
\end{definition}

It is straightforward to show that for any bounded set $T\subset\mathbb R^d$, $s(T)\leq d$ (see again \cite[Section 7.6]{vershynin2018high}). However, the stable dimension can be much lower than the algebraic dimension, even if the latter is allowed to be infinite. As we shall see, the stable dimension of the data support, appears in the upper bounds we derive for the generalisation error, and for the excess empirical error. We will also be using the following lemma about the relation of $\omega(\cdot)$ and $\psi(\cdot)$.
\begin{lemma}[{\cite[Section 7.6]{vershynin2018high}}]\label{th:equivalence}
    For any set $T\subset\mathbb R^d$, 
    \begin{equation}
        \omega(T-T)\leq\psi(T-T).
    \end{equation}\end{lemma}

The backbone of our two main results is an extension of the upper bound of the well-known Gordon's theorem \cite{gordon1988milman} (see also Theorem 5.6. in \cite{bandeira2015ten}), from the unit sphere to arbitrary sets. While we are aware of more general results that assume sub-gaussian random matrices (e.g. \cite[Section 9.1]{vershynin2018high}), we offer a simpler proof for the Gaussian case, that is free of any unspecified constants, and can thus be of interest in its own right. This is provided in the following lemma.

\begin{lemma}\label{th:Gordon's theorem generalisation}
    Let $R\in\mathbb R^{k\times d}$ be a matrix, with elements i.i.d. from $\mathcal N(0,1)$, and let $T\subset\mathbb R^d$ be a set, such that $\sup_{x\in T}\|x\|=b$. Also, let $a(k):=\E\|z_k\|$, where $z_k\sim\mathcal N(0_k,I_k)$. Then, for any $\epsilon>0$, with probability at least $1-\exp(-\epsilon^2/2b^2)$, we have
    \begin{equation}
        \sup_{x\in T}\|Rx\|\leq ba(k)+\omega(T)+\epsilon,
    \end{equation}
    where $\omega(\cdot)$ is the Gaussian width from Definition \ref{def:Gaussian width}.
\end{lemma}
It is well-known that $\frac k{\sqrt{k+1}}\leq a(k)\leq\sqrt k$. To prove Lemma \ref{th:Gordon's theorem generalisation}, we first recall a well-known inequalitiy regarding Gaussian processes (see \cite[Section 7.3]{vershynin2018high} and the references therein for the definitions and derivations).
\begin{lemma}[Sudakov-Fernique's inequality {\cite[Theorem 7.2.11]{vershynin2018high}}]\label{th:Sudakov-Fernique's inequality}
    Let $(X_t)_{t\in T}$ and $(Y_t)_{t\in T}$ be two mean-zero Gaussian processes and assume that for all $t,s\in T$, we have\begin{equation}\E(X_t-X_s)^2\leq\E(Y_t-Y_s)^2.\end{equation}Then, we have\begin{equation}\E\sup_{t\in T}X_t\leq\E\sup_{t\in T}Y_t.\end{equation}
\end{lemma}

\begin{IEEEproof}[Proof of Lemma \ref{th:Gordon's theorem generalisation}]
    We first define two mean-zero Gaussian processes as
    \begin{equation}
        X_{u,x}:=bg^\top u+h^\top x\qquad\text{ and }\qquad Y_{u,x}:=u^\top Rx
    \end{equation}
    where $(u,x)\in\mathcal S^{k-1}\times T$ and $g\sim\mathcal N(0_k,I_k)$ and $h\sim\mathcal N(0_d,I_d)$ are independent from each other.

    For all $(u,x),(u',x')\in\mathcal S^{k-1}\times T$, we have
    \begin{equation}
        \E(X_{u,x}-X_{u',x'})^2=2b^2-2b^2u^\top u'+\|x\|^2+\|x'\|^2-2x^\top x',
    \end{equation}
    and
    \begin{align}
        \E(Y_{u,x}-Y_{u',x'})^2&=\E(u^\top Rx-u'^\top Rx')^2\\
        &=\E\left(\sum_{i=1}^k\sum_{j=1}^d(R)_{ij}(u_ix_j-u'_ix'_j)\right)^2\\
        &=\sum_{i=1}^k\sum_{j=1}^d(u_ix_j-u'_ix'_j)^2\\
        &=\|ux^\top-u'x'^\top\|_F^2\\
        &=\tr((ux^\top-u'x'^\top)^\top (ux^\top-u'x'^\top))\\
        &=\|x\|^2+\|x'\|^2-2(u^\top u')(x^\top x').
    \end{align}
    
    Therefore, we find that
    \begin{equation}
        \E(X_{u,x}-X_{u',x'})^2-\E(Y_{u,x}-Y_{u',x'})^2=2(1-u^\top u')(b^2-x^\top x').
    \end{equation}
    
    This means that for all $(u,x),(u',x')\in\mathcal S^{k-1}\times T$ we have
    \begin{equation}
        \E(X_{u,x}-X_{u',x'})^2-\E(Y_{u,x}-Y_{u',x'})^2\geq0.
    \end{equation}
    
    Therefore, the conditions of Lemma \ref{th:Sudakov-Fernique's inequality} are satisfied, and thus we have
    \begin{equation}
        \E\sup_{(u,x)\in\mathcal S^{k-1}\times T}X_{u,x}\geq\E\sup_{(u,x)\in\mathcal S^{k-1}\times T}Y_{u,x}.
    \end{equation}
    
    Noting that
    \begin{equation}
        \E\sup_{u\in\mathcal S^{k-1}}\sup_{x\in T}X_{u,x}=ba(k)+\omega(T).
    \end{equation}
    and
    \begin{equation}
        \E\sup_{u\in\mathcal S^{k-1}}\sup_{x\in T}Y_{u,x}=\E\sup_{x\in T}\|Rx\|
    \end{equation}
    we conclude that
    \begin{equation}
        \E\sup_{x\in T}\|Rx\|\leq ba(k)+\omega(T).
    \end{equation}
    
    It remains to bound $\sup_{x\in T}\|Rx\|$ with high-probability away from its expectation. To this end, we claim that the function $f(R)=\sup_{x\in T}\|Rx\|$ is $b$-Lipschitz with respect to the Euclidean norm. To see why, let $R_1,R_2\in\mathbb R^{k\times d}$ be fixed matrices (which can also be seen as vectors in $\mathbb R^{kd}$), and note that
    \begin{align}
        \vert f(R_1)-f(R_2)\vert&=\vert\sup_{x\in T}\|R_1x\|-\sup_{x\in T}\|R_2x\|\vert\\
        &\leq\sup_{x\in T}\vert\|R_1x\|-\|R_2x\|\vert\\
        &\leq\sup_{x\in T}\|(R_1-R_2)x\|\\
        &\leq\sup_{x\in T}\|x\|\sigma_{\max}(R_1-R_2)\\
        &=b\sigma_{\max}(R_1-R_2)\\
        &\leq b\|R_1-R_2\|_F.
    \end{align}
    
    Invoking the upper bound of \cite[Theorem 2.26]{wainwright2019high}, we complete the proof.
\end{IEEEproof}
Applying Lemma \ref{th:Gordon's theorem generalisation}, we can derive the following uniform, high-probability upper bound for the generalisation error of the compressed hypothesis class.
\begin{theorem}[Compressed generalisation error]\label{th:generalisation error}
     Let $R\in\mathbb R^{k\times d}$, with elements i.i.d. from $\mathcal N(0,1/k)$, $T\subset(\mathcal X\times\mathcal Y)^2$ be the training set defined in \eqref{eq:training set}, $\mathcal M_k$ be the hypothesis class defined in \eqref{eq:compressed metric class}, $L^R_{\mathcal D}$ be the compressed true error defined in \eqref{eq:true error compressed}, and $\hat L^R_T$ be the compressed empirical error defined in \eqref{eq:empirical error compressed}. Then, for any $0<\epsilon<1$, and for all $M\in\mathcal M_k$, with probability at least $1-\epsilon$, we have
    \begin{multline}\label{eq:generalisation error}
        L^R_{\mathcal D}(M)-\hat L^R_T(M)\leq\\2\rho\sqrt{\frac kn}\left(1+\sqrt{\frac{s(\mathcal X)}k}+\sqrt{\frac{2\ln\frac2\epsilon}k}\right)^2+\sqrt{\frac{\ln\frac2\epsilon}{2n}}.
    \end{multline}
\end{theorem}
\begin{IEEEproof}
    Let $\mathcal P$ be a probability measure induced by the random variable $(X,Y)$, where $X:=(x,x')$ and $Y:=\I{y=y'}$, for $((x,y),(x',y'))\sim\mathcal D^2$. Also denote $\mathcal D_{\mathcal X}$ the marginal distribution induced by $\mathcal D$ on $\mathcal X$. Also let $\ell_{l,u}$ be the loss function defined in \eqref{eq:loss function}. Given a matrix $R$, we define the function class in the compressed space as
\begin{multline}
    \mathcal F_R=\{f_M:(x_1,x_2)\to\|M(Rx_1-Rx_2)\|^2\\:M\in\mathcal M_k\text{ and }x_1,x_2\in\mathcal X\}.
\end{multline}

Also, for all $i\in[n]$, let $X_i:=(x_{2i-1},x_{2i})$ and $Y_i:=\I{y_{2i-1},y_{2i}}$ be ``regrouped'' versions of the elements of $T$, defined in \eqref{eq:training set}. We are interested in upper bounding
\begin{equation}
    \sup_{f_M\in\mathcal F_R}\left(\E_{(X,Y)\sim\mathcal P}\ell_{l,u}(f_M(X),Y)-\frac1n\sum_{i=1}^n\ell_{l,u}(f_M(X_i),Y_i)\right)
\end{equation}

We then upper bound the Rademacher complexity\footnote{See Lemma \ref{th:Rademacher bound} for the definition of the Rademacher complexity} of $\mathcal F_R$, with respect to $\mathcal P$. Let $\sigma_1,\ldots,\sigma_n$ be  i.i.d. uniform $\{\pm1\}$-valued random variables. Modifying the proof of \cite[Theorem 1]{verma2015sample}, we obtain with probability at least $1-\epsilon$
{\fontsize{8}{8}\selectfont
\begin{align}
    &\mathcal R_{n,\mathcal D}(\mathcal F_R):=\frac1n\E_{\substack{\sigma_i,X_i\\i\in[n]}}\sup_{f_M\in\mathcal F_R}\sum_{i=1}^n\sigma_if_M(x_{2i-1},x_{2i})\\
    &=\frac1n\E_{\substack{\sigma_i,X_i\\i\in[n]}}\sup_{M\in\mathcal M_k}\sum_{i=1}^n\sigma_i(x_{2i-1}-x_{2i})^\top R^\top M^\top MR(x_{2i-1}-x_{2i})\\
    &\leq\frac1{\sqrt n}\sup_{M\in\mathcal M_k}\|M^\top M\|_F(\E_{(x,x')\sim\mathcal D_{\mathcal X}\times\mathcal D_{\mathcal X}}\|R(x-x')\|^4)^{1/2}\\
    &\leq\frac{\sqrt k}{\sqrt n\diam(\mathcal X)^2}(\E_{(x,x')\sim\mathcal D_{\mathcal X}\times\mathcal D_{\mathcal X}}\|R(x-x')\|^4)^{1/2}\\
    &\leq\frac{\sqrt k}{\sqrt n\diam(\mathcal X)^2}\sup_{x,x'\in\mathcal X}\|R(x-x')\|^2\\
    &\leq\frac{\sqrt k}{\sqrt n\diam(\mathcal X)^2}\left(\diam(\mathcal X)+\frac{\omega(\mathcal X-\mathcal X)}{\sqrt k}+\diam(\mathcal X)\sqrt{\frac{2\ln\frac1\epsilon}k}\right)^2\label{eq:use of lemma 1}\\
    &\leq\frac{\sqrt k}{\sqrt n\diam(\mathcal X)^2}\left(\diam(\mathcal X)+\frac{\psi(\mathcal X-\mathcal X)}{\sqrt k}+\diam(\mathcal X)\sqrt{\frac{2\ln\frac1\epsilon}k}\right)^2\label{eq:use of lemma 2}\\
    &=\sqrt{\frac kn}\left(1+\sqrt{\frac{s(\mathcal X)}k}+\sqrt{\frac{2\ln\frac1\epsilon}k}\right)^2.
\label{eq:Rademacher bound}\end{align}}

We used Lemma \ref{th:Gordon's theorem generalisation} to obtain \eqref{eq:use of lemma 1}, and the inequality of Lemma \ref{th:equivalence} to obtain \eqref{eq:use of lemma 2}. To complete the proof, we then invoke the well-known Rademacher bound, which we include for completeness in Lemma \ref{th:Rademacher bound}, combined with the union bound \cite[Theorem 1.2.11.b]{casella2002statistical}.
\begin{lemma}[Rademacher bound \cite{bartlett2002rademacher}]\label{th:Rademacher bound}
     Let $\mathcal D$ be a distribution over $\mathcal X\times\{0,1\}$ and let $\{(x_i,y_i)\}_{i=1}^n$ be a sample of size $n$ drawn i.i.d. from $\mathcal D$. Given a hypothesis class $\mathcal F$ and a loss function $\ell:\mathbb R\times\{0,1\}\to\mathbb R$ such that $\vert\ell(y',y)\vert\leq1$, for all $y',y\in\mathbb R$ and $\ell$ is $\rho$-Lipschitz in its first argument, then, for any $0<\epsilon<1$, with probability at least $1-\epsilon$ for all $f\in\mathcal F$, we have\small{\begin{equation}\E_{(x,y)\sim\mathcal D}\ell(f(x),y)\leq\frac1n\sum_{i=1}^n\ell(f(x_i),y_i)+2\rho\mathcal R_{n,\mathcal D}(\mathcal F)+\sqrt{\frac{\ln\frac1\epsilon}{2n}}\end{equation}}
     where $\mathcal R_{n,\mathcal D}(\mathcal F)$ is the \emph{Rademacher complexity} of the hypothesis class $\mathcal F$, given a sample of size $n$ i.i.d. from $\mathcal D$, and is defined as
     \begin{equation}\mathcal R_{n,\mathcal D}(\mathcal F):=\frac1n\E_\sigma\sup_{f\in\mathcal F}\sum_{i=1}^n\sigma_if(x_i)\end{equation}
     where $\sigma:=\begin{bmatrix}\sigma_1,\ldots,\sigma_n\end{bmatrix}^\top$ is a random vector, consisting of $n$, i.i.d., uniform, $\{\pm1\}$-valued random variables.
\end{lemma}
\end{IEEEproof}
We can see that the ambient dimension does not appear in the bound of Theorem \ref{th:generalisation error}, and is instead replaced by the stable dimension of the data support. This implies that, unless the data support fills the whole ambient space,  the empirical error calculated in the compressed space is closer to the true error in the compressed space.

The behaviour of this bound with $k$ and $n$ is as expected, since higher values of $k$ result in more complex hypothesis classes, whereas a larger $n$, reduces the discrepancy between the true end empirical error.
\begin{remark} For learning a Mahalanobis metric in the original data space, previous work of \cite{verma2015sample} implies the following uniform upper bound on the generalisation error.

For any $0<\epsilon<1$, and for all $M_0\in\mathcal M$, with probability at least $1-\epsilon$, we have
    \begin{equation} \label{eq:verma generalisation bound}
        L_{\mathcal D}(M_0)-\hat L_T(M_0)\leq2\rho\sqrt{\frac dn}+\sqrt{\frac{\ln\frac1\epsilon}{2n}},
   \end{equation}
    where $L_{\mathcal D}$ and $\hat L_T$ are respectively the true error defined in \eqref{eq:true error}, and the empirical error defined in \eqref{eq:empirical error}. If in addition a Frobenius norm constraint is imposed on the class \eqref{eq:metric class} (which we did not impose), then $d$ is replaced by the upper bound on the Frobenius norm constraint in the bound. Although our uniform bound for $\mathcal M_k$ in Theorem \ref{th:generalisation error} is similar in flavour to this latter result under the norm constraint, its purpose is different. In \cite{verma2015sample}, one tries to learn a metric with a low-Frobenius norm. In our case, we are instead interested to quantify the trade-off induced by the random projection between the generalisation error and the excess empirical error (see Theorem \ref{th:excess empirical error} below for the latter), without norm constraints. An advantage we gain is not having to know beforehand about the bounded Frobenius norm of the metric, instead we only need to set the projection dimension $k$. Besides this, of course, the main gain lies in the time and space savings of learning a $k\times k$ instead of a $d\times d$ matrix.
\end{remark}
However, a generalisation bound is not the complete story when we work with the RP-ed data, as there is usually a trade-off between accuracy and complexity. Intuitively, we can expect that, as the projection dimension $k$ decreases, we obtain a lower complexity of the compressed hypothesis class, but a higher empirical error (due to the potential distortion that results from the compression). We already upper bounded the former in Theorem \ref{th:generalisation error}, so we next upper bound the latter, with high-probability, as follows:
\begin{theorem}[Excess empirical error]\label{th:excess empirical error}
        Let $R\in\mathbb R^{k\times d}$, with elements i.i.d. from $\mathcal N(0,1/k)$, $T\subset(\mathcal X\times\mathcal Y)^2$ be the training set defined in \eqref{eq:training set}, $\mathcal M$ and $\mathcal M_k$ be the hypothesis classes defined in \eqref{eq:metric class} and \eqref{eq:compressed metric class} respectively, $\hat L_T$ be the empirical error defined in \eqref{eq:empirical error}, and $\hat L^R_T$ be the compressed empirical error defined in \eqref{eq:empirical error compressed}. Then, for any $0<\epsilon<1$, and for all $M\in\mathcal M_k$ and $M_0\in\mathcal M$, with probability at least $1-\epsilon$, we have
    \begin{equation}\label{eq:excess empirical error}
    \hat L^R_T(M)-\hat L_T(M_0)\leq\rho\left(1+\sqrt{\frac{s(\mathcal X)}k}+\sqrt{\frac{2\ln\frac1\epsilon}k}\right)^2
\end{equation}
\end{theorem}
\begin{IEEEproof}
    Consider any pair of hypotheses, $M\in\mathcal M_k$, and $M_0\in\mathcal M$. Using the $\rho$-Lipschitz property of $\ell_{l,u}$, we have
\begin{multline}\label{eq:empirical bound}
    \hat L^R_T(M)-\hat L_T(M_0)\leq\\\frac\rho n\sum_{i=1}^n\vert\|MR(x_{2i-1}-x_{2i})\|^2-\|M_0(x_{2i-1}-x_{2i})\|^2\vert.
\end{multline}

To upper bound the absolute value in \eqref{eq:empirical bound}, we need to both lower and upper bound the quantity inside, with respect to $R$, and take the maximum of the two. There are two terms inside the maximum, which must be lower and upper bounded separately.

For the first term, recall that $\sigma_{\max}(M)=\sigma_{\max}(M_0)=1/\diam(\mathcal X)$ by their definition. We invoke Lemma \ref{th:equivalence} and both bounds of Lemma \ref{th:Gordon's theorem generalisation} to obtain our results. For the upper bound, with probability at least $1-\epsilon$, we have for all $i\in[n]$
\begin{align}
    \|MR(x_{2i-1}-x_{2i})\|&\leq\sigma_{\max}(M)\|R(x_{2i-1}-x_{2i})\|\\
    &\leq\frac1{\diam(\mathcal X)}\|R(x_{2i-1}-x_{2i})\|\\
    &\leq\frac1{\diam(\mathcal X)}\sup_{x,x'\in\mathcal X}\|R(x-x')\|\\
    &\leq1+\sqrt{\frac{s(\mathcal X)}k}+\sqrt{\frac{2\ln\frac1\epsilon}k}.
\end{align}

For the lower bound, with probability at least $1-\epsilon$, we have for all $i\in[n]$
\begin{equation}
    \|MR(x_{2i-1}-x_{2i})\|\geq0.
\end{equation}

For the second term, since $\sigma_{\max}(M_0)=1/\diam(\mathcal X)$, we have for all $i\in[n]$
\begin{equation}
    0\leq\|M_0(x_{2i-1}-x_{2i})\|\leq1.
\end{equation}

Plugging the lower and upper bounds into \eqref{eq:empirical bound}, we obtain, with probability at least $1-\epsilon$
{\small
\begin{equation}
    \hat L^R_T(M)-\hat L_T(M_0)\leq\max\left\{\rho\left(1+\sqrt{\frac{s(\mathcal X)}k}+\sqrt{\frac{2\ln\frac1\epsilon}k}\right)^2,\rho\right\}.
\end{equation}
}

Since the first term inside the maximum is always greater than $\rho$, this simplifies to our desired result.
\end{IEEEproof}
Examining the bound in Theorem \ref{th:excess empirical error}, we can see it does not depend on the ambient dimension, but on the stable dimension of the data support, just like the bound in Theorem \ref{th:generalisation error}. This means that if the empirical error in the ambient space is small, the empirical error in the compressed space scales with the stable dimension, instead of the ambient dimension. It is also decreasing in $k$ as expected. Finally, the sample size, $n$, does not appear at all, as it is assumed the same for training both $M$ and $M_0$, and simplifies out in the derivation.

\begin{remark}
     The motivation behind generalising Gordon's theorem to our Lemma \ref{th:Gordon's theorem generalisation}, was to make our main results dimension-free. Indeed, applying the original Gordon's theorem to our derivations of Theorems \ref{th:generalisation error} and \ref{th:excess empirical error}, we would obtain the same formulas, but with $d$ in place of $s(\mathcal X)$. As we already mentioned, it can be the case that $s(\mathcal X)\ll d$, thus our results adapt to a notion of low-ID, and unveiling such a low-ID dependence, was the overall goal of our paper.
\end{remark}

To summarise, a Gaussian random projection incurs a lower generalisation gap for Mahalanobis metric learning, but induces an excess empirical error, compared to learning the metric in the ambient space. In our bounds, both quantities depend on the stable dimension of the data support, instead of the ambient dimension, so these bounds automatically tighten when the stable dimension is low. We next illustrate the effects that the stable dimension has on metric learning, in numerical experiments.
\section{Experimental results}\label{sec:experimental results}
In this section we conduct numerical experiments to validate our theoretical guarantees in practice, on both synthetic and benchmark data sets, when learning a Mahalanobis metric in compressed settings. To design our experiments, let us recall that we derived theoretical guarantees for two quantities:
\begin{itemize}
    \item 
    the generalisation error of metric learning under Gaussian random projection; and
    \item 
    the excess empirical error incurred relative to that of metric learning in the ambient space;
\end{itemize}
and that, both of them, were found to depend on $k$ and $s(\mathcal X)$, instead of $d$.

The main goal of our experiments, is to find how much distortion is incurred by different choices of the projection dimension, $k$, and how is it affected by $s(\mathcal X)$. The motivation is that if the distortion is minimal for some $k$, we can enjoy almost the same empirical performance as in the ambient space, but with a much lower time complexity, as we operate in the compressed space. Therefore, the trade-off between accuracy and complexity can be minimised, by choosing an appropriate value for $k$. Due to space constraints, in our figures, we only report the error rates achieved by the compressive algorithm, and omit the computational time -- which is clearly strictly increasing in $k$.

We start with a brief overview of our experimental setup. We first choose the original data set in the ambient space. We then perform a Gaussian random projection and train a metric using Large Margin Nearest Neighbour (LMNN) \cite{weinberger2009distance} in the compressed space. Finally, we use $1$-Nearest Neighbours ($1$-NN) to evaluate the quality of the learned Mahalanobis metric on the compressed set, and report the out-of-sample test error. We repeat this process 10 times independently, for a number of choices of the projection dimension. As in \cite{verma2015sample}, we opt for using $1$-NN to ``normalise'' the metric error to the interval $[0,1]$, and allow for easier comparisons, which would be trickier with the metric loss from \cite{weinberger2009distance}.

\subsection{Experiments with synthetic sets}\label{subsec:synthetic experiments}
Synthetic data allow easy control of the stable dimension of their support, hence they allow us to test the explanatory abilities of our theoretical results. We take the data support to be an ellipsoid of the form $\mathcal X=A\mathcal S^{d-1}$, where $A\in\mathbb R^{d\times d}$ with $\sigma_{\max}(A)=1$ is a diagonal, positive-definite matrix (without loss of generality, since the algorithm is rotation-invariant). The stable dimension of the support is determined by the eigenvalues of $A$. Preliminary experimentation, has shown that the rate of decay of the eigenvalues of the ellipsoid, affects the error. We therefore consider different rates of decay of the eigenvalues of $A$. We generate a sample set of $2000$ instances, $\{x_i\}_{i=1}^{2000}$, sampled uniformly randomly over $\mathcal X$, and employ a train/test ratio of $80\%/20\%$.

By construction, in this setting the stable dimension of $\mathcal X$ has the closed form expression $s(\mathcal X)=(\|A\|_F/\sigma_{\max}(A))^2$ \cite[Section 7.6]{vershynin2018high}. Hence, according to our theoretical results, we expect that increasing $d$ should not blow up the out of sample test error, as long as $s(\mathcal X)$ does not increase significantly. We employ the Gaussian random projection in these experiments, as studied in our theory.

We want to compare the out-of-sample test error in the compressed space, with the error in the ambient space, across several choices of $k$. Due to this, we consider settings where the empirical error in the ambient space is small, and thus it is enough to examine only the empirical error in the compressed space, thus saving computational time. For the purpose of maintaining a small (but not zero) empirical error in the ambient space, we considered linearly separable class supports, where 1-NN can achieve almost perfect classification. Specifically, the original labels were set to $y_i:=\sign(w^\top x_i)$ for all $i\in[2000]$, where $w$ was sampled from $\mathcal N(0_d,I_d)$, and then fixed for each value of $d$. To combat randomness, we fixed a sequence of $1000$ elements, sampled i.i.d. from $\mathcal N(0,1)$, and for each $d$, we used the first $d$ elements of this sequence, as coordinates for $w$.
\begin{figure*}[!ht]
    \centering
    \includegraphics{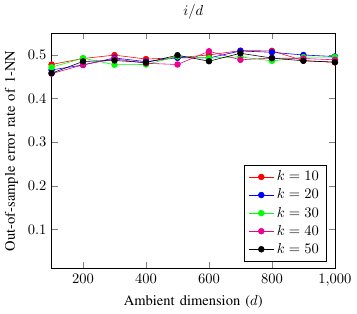}
    \includegraphics{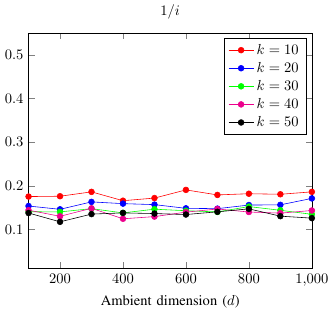}
    \includegraphics{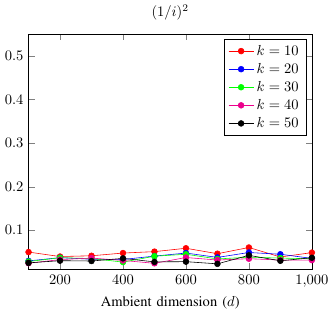}
    \caption{Out-of-sample error of $1$-NN on compressed synthetic data sets, with metric learning, averaged over $10$ Gaussian random projections, for several choices of $d$. The data support was $\mathcal X=A{\mathcal S}^{d-1}$, where $A\in\mathbb{R}^{d\times d}$ is a diagonal matrix, and the titles of the subplots shows its $i$-th diagonal element, for $i\in[d]$. The legends show the projection dimension, $k$.}
    \label{fig:empirical error synthetic}
\end{figure*}

Figure \ref{fig:empirical error synthetic} shows the empirical results obtained with metric learning. As expected from the theory, we see that the error is affected by the stable dimension, which, in turn, depends on the rate of decay on the eigenvalues of $A$ (shown in the legends), and is unaffected by the ambient dimension. To confirm, we repeated these experiments with different values of $d$, and for different decay rates on the eigenvalues of $A$. For small decay rates (left sub-figure), the stable dimension increases rapidly with $d$, and the error-rate is close to $0.5$, as in a random guess. For larger decay rates (middle and right sub-figures) the stable dimension increases slowly, and the error-rate is much lower.


\subsection{Experiments with benchmark data sets}\label{subsec:benchmark experiments}
Benchmark data sets will serve to test the usefulness and effectiveness of metric learning under compression in a more general context, and its adaptability to noisy settings. In real data, the value of the stable dimension of the support is unknown, but one may expect some structure that metric learning can exploit. We follow the same experimental protocol as in synthetic data sets (80\%/20\% split), and compute the empirical error, for varying degrees of compression. We want to test if the trade-off can be minimised by some value of $k$.

Our test experiments are somewhat inspired from the evaluation idea in \cite{verma2015sample}, where noise features were appended to low-dimensional data to test the abilities of metric learning. 
We start from three benchmark UCI data sets with moderate ambient dimension from \cite{dua2019uci}: \textsc{Ionosphere} (2 labels, 33 features, 351 instances), \textsc{Wine} (3 labels, 13 features, 178 instances), and \textsc{Sonar} (2 labels, 60 features, 208 instances). For each set, we normalised its features to $[0,1]$, embedded it onto a higher-dimensional ambient space, and added some Gaussian noise to all features and all instances, with variance $\gamma$. This simulates the ``noisy subspace hypothesis'', in which the data cluster in a noisy low-dimensional subspace.

We aim to test whether Gaussian random projection is still able to preserve information from the features that span the underlying subspace. We also repeated the experiments for different values of $\gamma$, to test how easily metric learning can adapt in each case.
\begin{figure*}[!ht]
    \centering
    \includegraphics{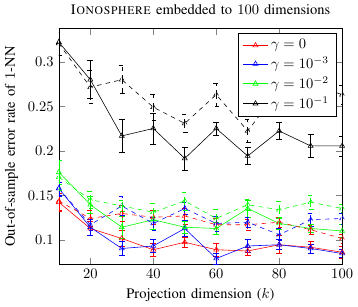}
    \includegraphics{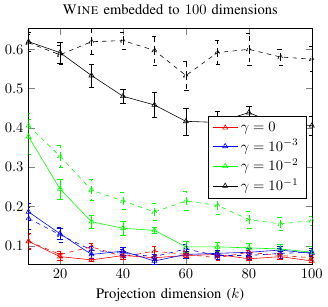}
    \includegraphics{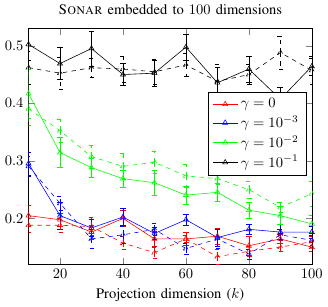}
    \caption{Out-of-sample error of $1$-NN classification with metric learning (solid lines) and with Euclidean metric (dashed lines), of benchmark UCI data sets. All data sets were normalised to $[0,1]$, embedded to a $100$-dimensions, and had i.i.d. Gaussian random noise of variance $\gamma$ (shown in the legend) added to each of their instances. A train/test ratio of $80\%/20\%$ was used. The curves represent averages over $10$ independent Gaussian random projections. The error bars show intervals of one standard error.}
    \label{fig:empirical error real}
\end{figure*}
Figure \ref{fig:empirical error real} shows the results. As we can see, the higher the noise variance $\gamma$, the higher the average error incurred by the algorithm. However, in almost all cases, there seems to be a lower bound for $k$, above which the performance stops increasing significantly. This means that the trade-off between accuracy and complexity can be minimised, by choosing that value of $k$ (e.g. by employing cross-validation type procedures).

Regarding the performance of metric learning, compared to the Euclidean metric, es expected, it depends on the unknown structure of the data and the available sample size, although in the higher-noise regime we see a consistently outperformance from learning the metric.

\section{Conclusions and Future Work}\label{sec:conclusions and future work}
We considered Mahalanobis metric learning when working with a randomly compressed version of the data. We derived high-probability theoretical guarantees for its generalisation error, as well as for its excess empirical error under Gaussian random projection. We showed theoretically that both quantities are unaffected by the ambient dimension, and instead depend on the stable dimension of the data support. We supported these findings with experiments on both synthetic and benchmark data sets in conjunction with Nearest Neighbour classification, using its empirical performance to evaluate the learnt metric learning.

In this work we only considered properties of the support of the data. Future work may focus on effects from other distributional traits. This may be particularly useful in settings where the covariance of the distribution is far from isotropic, and the data support is only bounded with high-probability. Related work has been done for quadratic classifiers in \cite{latorre2021effect}, which showed that the effective rank of the second-moment matrix (a measure of ID) affects the generalisation error. The second-moment matrix is usually unknown, so it would be insightful to see how metric learning can automatically adapt to some particular structure in that matrix.

Another possible extension is to study the setting where each compressed instance is perturbed by random noise. Metric learning under noisy regimes has already been examined, e.g. \cite{lim2013robust}, but only in the ambient space. Considering the effect of noise on metric learning under compression may also be of interest in many real-world settings.


\bibliographystyle{unsrt}
\bibliography{references.bib}

\end{document}